\documentclass[conference,english]{IEEEtran}
\IEEEoverridecommandlockouts

\usepackage{cite}
 \usepackage{amsmath,amssymb,amsfonts,amsthm}
\usepackage[pdftex]{graphicx}
\usepackage{textcomp}
\usepackage{xcolor}
\usepackage{booktabs}
\usepackage{multirow}
\usepackage{multicol}
\usepackage{graphicx}
\usepackage[T1]{fontenc}
\usepackage[utf8]{inputenc}
\usepackage{babel}
\usepackage{mathtools}
\usepackage{algorithm}
\usepackage{tikz}
\usepackage{pgfplots}
\usepackage{glossaries-extra}
\usepackage{siunitx}
\usepackage{algpseudocodex}
\usepackage{svg}
\usepackage{etoolbox}
\usepackage{balance}
\usepackage{microtype}
\usepackage{mathtools}
\usepackage[hidelinks]{hyperref}
\usepackage[capitalise]{cleveref}
\usepackage{xcolor, soul}
\sethlcolor{red}

\setabbreviationstyle[acronym]{long-short}

\svgsetup{inkscapelatex=false}

\pgfplotsset{compat=newest,
every axis/.style={
    grid=major,
    xlabel near ticks,
    ylabel near ticks,
    legend pos=south east,
    legend style={font=\footnotesize},
    every x tick label/.append style = {font=\footnotesize},
    every y tick label/.append style = {font=\footnotesize},
    cycle list/Set2-8,
    cycle multiindex* list={
    mark list\nextlist
    Set2-8\nextlist
    linestyles\nextlist
    },
    every axis plot/.append style={mark size=1pt, thick},
    enlarge y limits=0.025,
    enlarge x limits=0
}
}
\graphicspath{{figures/}}

\usepgfplotslibrary{colormaps}
\usepgfplotslibrary{colorbrewer}
\usepgfplotslibrary{external}

\makeatletter
\newcommand\fs@betterruled{%
  \def\@fs@cfont{\bfseries}\let\@fs@capt\floatc@ruled
  \def\@fs@pre{\vspace*{5pt}\hrule height.8pt depth0pt \kern2pt}%
  \def\@fs@post{\kern2pt\hrule\relax}%
  \def\@fs@mid{\kern2pt\hrule\kern2pt}%
  \let\@fs@iftopcapt\iftrue}
\floatstyle{betterruled}
\restylefloat{algorithm}
\makeatother

\usepackage{tikz}
\usetikzlibrary{backgrounds, calc, chains, fit, matrix, positioning, shadows, shapes, intersections, circuits.ee}
\tikzset{input/.style={}}
\tikzset{output/.style={}}
\tikzset{op/.style={circle, draw, fill=black!10, minimum size=2.5ex, inner sep=0ex}}
\tikzset{filter/.style={rectangle, draw, thick, fill=black!10, minimum size=3.5ex, inner sep=1ex}}
\tikzset{nn/.style={trapezium, trapezium angle=80, draw, thick, fill=black!10, inner sep=1ex}}
\tikzset{branch/.style={circle, draw, thick, fill=black, minimum size=.5ex, inner sep=0ex}}
\tikzset{tensor/.style={rectangle, draw, fill=white, minimum size=2em, double copy shadow={shadow xshift=.5ex,shadow yshift=-.5ex}}}
\tikzset{rounded/.style={rounded rectangle, draw, thick, fill=black!10, minimum size=3.5ex, inner xsep=1ex}}
\tikzset{image/.style={rectangle, draw, fill=white, minimum size=2em}}
\tikzset{>=direction ee}
\tikzset{/tikz/thin/.style={line width=.9pt}}
\tikzset{/tikz/thick/.style={line width=1.4pt}}
\tikzset{every path/.style={thin}}

\usepackage[none]{hyphenat}
\renewcommand{\topfraction}{.9}
\renewcommand{\bottomfraction}{.9}
\renewcommand{\dbltopfraction}{.9}

\renewcommand{\textfraction}{0}
\renewcommand{\floatpagefraction}{.8}
\renewcommand{\dblfloatpagefraction}{.8}

\allowdisplaybreaks[2]
\usepackage{caption}
\usepackage{subfigure}

\renewcommand{\vec}[1]{\mathbf{#1}}
\newcommand{\vecs}[1]{\boldsymbol{#1}}

\newcommand{\nv}{\vec{n}}

\newcommand{\sv}{\vec{s}}

\newcommand{\xv}{\vec{x}}
\newcommand{\yv}{\vec{y}}
\newcommand{\zv}{\vec{z}}

\newcommand{\Thetav}{\vecs{\Theta}}
\newcommand{\Phiv}{\vecs{\Phi}}

\newcommand{\Cc}{{\cal C}}
\newcommand{\Dc}{{\cal D}}

\newcommand{\Lc}{{\cal L}}

\newcommand{\Nc}{{\cal N}}

\newcommand{\Pc}{{\cal P}}

\newcommand{\CC}{\mathbb{C}}

\newcommand{\NN}{\mathbb{N}}
\newcommand{\RR}{\mathbb{R}}
\newcommand{\II}{\mathbb{I}}

\newcommand{\LB}{\left(}
\newcommand{\RB}{\right)}

\newcommand{\LSB}{\left[}
\newcommand{\RSB}{\right]}

\newcommand{\EE}{{\mathbb{E}}}

\newcommand{\argmin}[1]{\underset{#1}{\operatorname{arg}\,\operatorname{min}}\;}

\newcommand\norm[1]{\left\lVert#1\right\rVert}

\newcommand{\card}[1]{\vert{#1}\vert}
\newcommand{\logn}[2]{\mathop{\mathrm{log}_{#1} \LB #2\RB}}

\newcommand{\Pavg}{P_\mathrm{avg}}

\newcommand{\Dctrain}{\Dc_\mathrm{train}}

\newcommand{\removed}[1]{}

\newacronym{ACM}{ACM}{adaptive coding and modulation}
\newacronym{ADC}{ADC}{analog-to-digital conversion}
\newacronym{AGC}{AGC}{automatic gain control}
\newacronym{AWGN}{AWGN}{additive white Gaussian noise}
\newacronym{BER}{BER}{bit error rate}
\newacronym{BLER}{BLER}{block error rate}
\newacronym{BP}{BP}{backpropagation}
\newacronym{BPTT}{BPTT}{backpropagation through time}
\newacronym{CE}{CE}{cross-entropy}
\newacronym{CFO}{CFO}{carrier frequency offset}
\newacronym{CSI}{CSI}{channel state information}
\newacronym{DAC}{DAC}{digital-to-analog conversion}
\newacronym{DL}{DL}{deep learning}
\newacronym{DFT}{DFT}{discrete Fourier transform}
\newacronym{FFT}{FFT}{fast Fourier transform}
\newacronym{GAN}{GAN}{generative adversarial network}
\newacronym{GRU}{GRU}{gated recurrent unit}
\newacronym{iid}{i.i.d.\@}{independent and identically distributed}
\newacronym{IFFT}{IFFT}{inverse fast Fourier transform}
\newacronym{KL}{KL}{Kullback-Leibler}
\newacronym{LSTM}{LSTM}{long short-term memory}
\newacronym{MDP}{MDP}{Markov decision process}
\newacronym{ML}{ML}{machine learning}
\newacronym{MLP}{MLP}{multilayer perceptron}
\newacronym{MIMO}{MIMO}{multiple-input multiple-output}
\newacronym{MSE}{MSE}{mean squared error}
\newacronym{NN}{NN}{neural network}
\newacronym{DNN}{DNN}{deep neural network}
\newacronym{OFDM}{OFDM}{orthogonal frequency-division multiplexing}
\newacronym{pdf}{pdf}{probability density function}
\newacronym{pmf}{pmf}{probability mass function}
\newacronym{PSNR}{PSNR}{peak signal to noise ratio}
\newacronym{RBF}{RBF}{Rayleigh block-fading}
\newacronym{ReLU}{ReLU}{rectified linear unit}
\newacronym{RL}{RL}{reinforcement learning}
\newacronym{RNN}{RNN}{recurrent neural network}
\newacronym{SFO}{SFO}{sampling frequency offset}
\newacronym{SNR}{SNR}{signal-to-noise ratio}
\newacronym{SINR}{SINR}{signal-to-interference-plus-noise ratio}
\newacronym{SGD}{SGD}{stochastic gradient descent}
\newacronym{wrt}{w.r.t.\@}{with respect to}

\newacronym{OAC}{OAC}{over-the-air computation}
\newacronym{MAC}{MAC}{multiple access channel}
\newacronym{SIC}{SIC}{successive interference cancellation}
\newacronym{TDMA}{TDMA}{time division multiple access}
\newacronym{NOMA}{NOMA}{non-orthogonal multiple access}
\newacronym{CL}{CL}{curriculum learning}
\newacronym{JSCC}{JSCC}{joint source-channel coding}
\newacronym{DeepJSCC}{DeepJSCC}{deep joint source-channel coding}
\newacronym{MTL}{MTL}{multi-task learning}
\newacronym{MIL}{MIL}{multi-instance learning}
\newacronym{DML}{DML}{deep metric learning}
\newacronym{IoT}{IoT}{Internet of Things}
\newacronym{SSIM}{SSIM}{structural similarity index measure}
\newacronym{MS-SSIM}{MS-SSIM}{multi-scale \gls{SSIM}}
\newacronym{DDPM}{DDPM}{denoising diffusion probabilistic models}
\newacronym{MVL}{MVL}{multi-view learning}
\newacronym{CNN}{CNN}{convolutional neural network}
\newacronym{LPIPS}{LPIPS}{learned perceptual image patch similarity}
\newacronym{BPG}{BPG}{Better Portable Graphics}

\newacronym{IoE}{IoE}{Internet of everything}

\newacronym{V2X}{V2X}{vehicle-to-everything}

\newacronym{AR/VR}{AR/VR}{augmented/virtual reality}

\newacronym{DSC}{DSC}{distributed source coding}

\newacronym{ANN}{ANN}{artificial neural network}

\newacronym{BCR}{BCR}{bandwidth compression ratio}
\newacronym{BR}{BR}{bandwidth ratio}
\newacronym{OPTA}{OPTA}{Optimal Performance Theoretically Attainable}
\newacronym{AF}{AF}{attention feature}

\begin{document}

\title{Distributed Deep Joint Source-Channel Coding 
with Decoder-Only Side Information\\
\thanks{The present work has received funding from the European Union’s Horizon 2020 Marie Skłodowska Curie Innovative Training Network Greenedge (GA. No. 953775), from CHIST-ERA project SONATA (CHIST-ERA-20-SICT-004) funded by EPSRC-EP/W035960/1, and from NYU Wireless. For the purpose of open access, the authors have applied a Creative Commons Attribution (CC BY) license to any Author Accepted Manuscript version arising from this submission.}
}

\author{\IEEEauthorblockN{Selim F. Yilmaz\IEEEauthorrefmark{2}, Ezgi Ozyilkan\IEEEauthorrefmark{3}, Deniz Gündüz\IEEEauthorrefmark{2}, Elza Erkip\IEEEauthorrefmark{3}}
\IEEEauthorblockA{\IEEEauthorrefmark{2}Department of Electrical and Electronic Engineering, Imperial College London, UK \\ \{s.yilmaz21, d.gunduz\}@imperial.ac.uk   \\ 
\IEEEauthorrefmark{3}Department of Electrical and Computer Engineering, New York University, USA \\ \{ezgi.ozyilkan, elza\}@nyu.edu}  
}


\setcounter{topnumber}{20}
\setcounter{bottomnumber}{20}
\setcounter{totalnumber}{20}

\renewcommand{\topfraction}{.9}
\renewcommand{\bottomfraction}{.9}
\renewcommand{\dbltopfraction}{.9}

\renewcommand{\textfraction}{0}
\renewcommand{\floatpagefraction}{.8}
\renewcommand{\dblfloatpagefraction}{.8}

\addtolength{\textfloatsep}{-0.5em}

\clubpenalty=0
\widowpenalty=0
\displaywidowpenalty=0

\abovedisplayskip=1ex
\abovedisplayshortskip=1ex
\belowdisplayskip=1ex
\belowdisplayshortskip=1ex

\allowdisplaybreaks[2]
\interdisplaylinepenalty=2500

\maketitle

\begin{abstract}
We consider low-latency image transmission over a noisy wireless channel when correlated side information is present only at the receiver side (the Wyner-Ziv scenario). In particular, we are interested in developing practical schemes using a data-driven \gls{JSCC} approach, which has been previously shown to outperform conventional separation-based approaches in the practical finite blocklength regimes, and to provide graceful degradation with channel quality. We propose a novel neural network architecture that incorporates the decoder-only side information at multiple stages at the receiver side. Our results demonstrate that the proposed method succeeds in integrating the side information, yielding improved performance at all channel conditions in terms of the various quality measures considered here, especially at low channel \glspl{SNR} and small \glspl{BR}. We have made the source code of the proposed method public to enable further research, and the reproducibility of the results.
\end{abstract}




\glsresetall

\section{Introduction}


Conventional communication systems follow a two-step approach for the transmission of image/video data: (i) the source coding stage eliminates inherent redundancy within the image, and (ii) the channel coding stage introduces structured redundancy with error correcting codes to enable resiliency against channel's corrupting effects, such as noise and fading. Although Shannon's \emph{separation theorem}~\cite{shannon} proves that such a modular two-step source and channel coding approach is theoretically optimal in the asymptotic limit of infinitely long source and channel blocklengths, it is known that the optimality of separation no longer holds in the finite blocklength, non-ergodic or multi-user scenarios. Such scenarios arise in many time-critical emerging applications, such as \gls{IoE}, \gls{V2X}, as well as \gls{AR/VR} applications.

\begin{figure}
\includegraphics[width=1\linewidth]{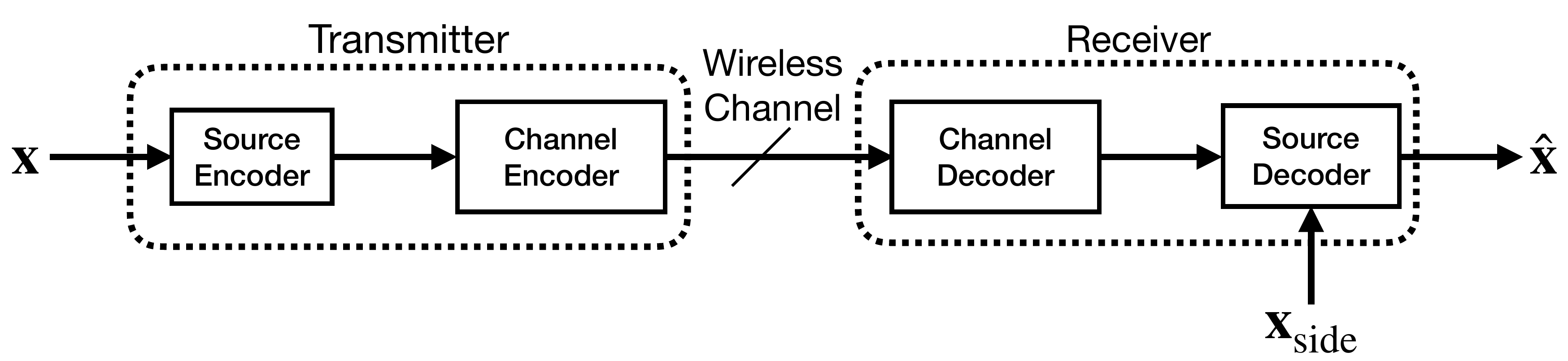}
\includegraphics[width=1\linewidth]{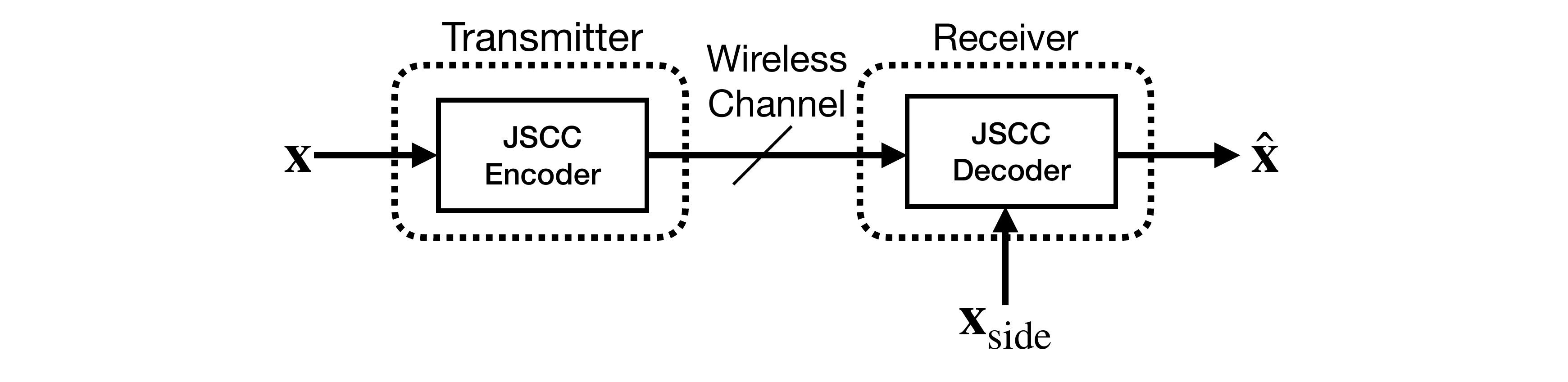}
\caption{Separation-based (\textbf{top}) vs. JSCC-based (\textbf{bottom}) communication schemes, having decoder-only side information.}
\label{fig:sys}
\end{figure} 

Despite having been investigated from a theoretical standpoint for many decades~\cite{1057373, 1057376}, \gls{JSCC} schemes found limited use in practical applications, mainly due to their high complexity and the difficulty in designing such codes. Recently, a \gls{DNN}-based \gls{JSCC} scheme \cite{bourtsoulatze2019deep}, namely \gls{DeepJSCC}, has achieved remarkable results and rekindled research interest in this direction. Specifically, the authors propose to use a data-driven approach to learn nonlinear mappings directly from the input image space to the channel input symbols in an end-to-end fashion, by adopting an autoencoder architecture. \gls{DeepJSCC} approach not only enjoys improved performance at a specific known channel state, but also exhibits \emph{graceful degradation} with channel \gls{SNR}, unlike the separation-based schemes that suffer from the \emph{cliff effect} phenomenon; that is, their performance collapses when the channel SNR falls below a certain threshold. 

In this paper, we are interested in the scenario, in which the receiver has access to a correlated \emph{side information} sequence (see Fig.~\ref{fig:sys}). For example, consider a distributed network of cameras, which aim to transmit their images to a joint central processing unit. In such a scenario, the images are highly correlated; hence, transmitting each image independently results in a significant communication overhead that scales with the number of cameras. In this work, as a first step towards exploiting source correlations in a large network, we consider the simple scenario with two cameras, where the first camera wants to send its image to the second camera.

It is known that separate source and channel coding  remains to be optimal in the case of communication with decoder-only side information~\cite{sys_jscc}. This is achieved by Wyner-Ziv lossy compression of the source sequence exploiting decoder's side information through binning~\cite{1055508}, followed by capacity-achieving channel coding. However, the price to pay for near-optimal performance is high complexity and delay since achieving the channel capacity and the Wyner-Ziv rate--distortion function necessitates large blocklengths. We instead leverage the universal function approximation capability of neural networks~\cite{hornik_et_al} to find constructive solutions for JSCC with decoder-only side information  in the non-asymptotic regime.


Our goal is to design a practical JSCC scheme that can benefit from the side information at the receiver. Our main contributions can be summarized as follows:
\begin{itemize}
    \item To the best of our knowledge, we introduce the first \gls{DeepJSCC}-based image transmission method that explicitly exploits decoder-only side information, termed \emph{DeepJSCC-WZ}. The proposed transmission scheme is an important building block towards fully distributed practical \gls{DeepJSCC} schemes for correlated image/video signals.
    \item We demonstrate that our method significantly outperforms both the point-to-point \gls{DeepJSCC} scheme (with no side information) and the separation-based scheme with side information, in terms of both traditional and perception-oriented fidelity metrics for all the considered channel \gls{SNR} and \gls{BR} values.
    \item As an upper bound, we also provide a solution for the scenario in which the side information is available at both the encoder and decoder. This allows us to quantify the performance gain by providing the side information also to the encoder.
    \item To facilitate further research and reproducibility, we also provide the source code of our framework and simulations on \href{https://github.com/ipc-lab/deepjscc-wz}{\texttt{github.com/ipc-lab/deepjscc-wz}}.
\end{itemize}

\section{Related Works}
\label{sec:related_works}

In this section, we briefly go over previous works upon which we construct our methods.

\subsection{Distributed Source Coding (DSC)}

Notable prior work on the source coding side is related to the \emph{distributed source coding} (DSC) problems. Slepian and Wolf~\cite{1055037} proved their seminal result that an encoder that does not observe a correlated side information can asymptotically achieve the same compression rate as the one that does (in both cases, the decoder has access to the side information), if the joint distribution statistics are known and compression is lossless. Later, Wyner and Ziv~\cite{1055508} characterized the rate--distortion function with side information available at both the encoder and decoder, or only at the decoder. Surprisingly, they showed that there is no rate loss in the latter scenario compared to the former, if the sources are Gaussian and mean-squared error is set as the distortion criterion. Practical research effort for the Wyner-Ziv setup has been spearheaded by distributed source coding using syndromes (DISCUS)~\cite{DISCUS}, which formulated the compression problem as a dual source-channel coding problem.

Recently, distributed deep neural compression schemes have been proposed in~\cite{mital2022neural, mital2023neural, ozyilkan2023learned, zhang2023ldmic}. These works, inspired by the Slepian-Wolf and Wyner-Ziv results~\cite{1055037, 1055508}, are concerned with exploiting the side information in order to further compress the primary input source, compared to the point-to-point (having no side information) compression scenario. In particular, the recent work in~\cite{ozyilkan2023learned} demonstrated that neural Wyner-Ziv compressors can learn ``random binning" structures, akin to the achievability part of the Wyner-Ziv theorem. However, these works do not consider the impact of the wireless channel.

\subsection{Deep Joint Source-Channel Coding (DeepJSCC)}

The first work employing DNN-based JSCC approach in wireless image transmission, termed \gls{DeepJSCC}, was originally proposed in~\cite{bourtsoulatze2019deep}. This data-driven communication scheme has later been extended to different channel models~\cite{9714510, yilmaz2023distributed}, different source signals~\cite{tung2021deepwive, han2022semanticpreserved}, inference problems~\cite{jankowski}, multi-user scenarios~\cite{yilmaz2023distributed,bian2022deep} and as well as to perceptual quality-oriented image transmission~\cite{erdemir2022generative,yilmaz2023high}. The authors in~\cite{9954060} proposed data-driven point-to-point and distributed JSCC schemes, employing sinusoidal representation networks (SIRENs) that are inspired by the Shannon-Kotel'nikov mappings~\cite{4768576} for the transmission of \gls{iid} and multivariate Gaussian sources over orthogonal Gaussian channels. However, for the decoder-only side information case, their approach only considers synthetic datasets and specific correlation patterns, and it is unclear how this approach would scale up to realistic and high-dimensional correlated information sources, such as stereo images, which we consider in this paper. 

\textbf{Notation:} Unless stated otherwise; boldface lowercase letters denote tensors (e.g., $\vec{p}$), non-boldface letters denote scalars (e.g., $p$ or $P$), and uppercase calligraphic letters denote sets (e.g., $\Pc$). $\RR$, $\NN$, $\CC$ denote the set of real, natural and complex numbers, respectively. $\card{\Pc}$ denotes the cardinality of set $\Pc$. We define $[n]\triangleq\{1,2,\cdots,n\}$, where $n\in\NN^+$, and $\II \triangleq [255]$.

\section{Problem Formulation}
\label{sec:system_model}

We consider the wireless image transmission problem over an \gls{AWGN} channel, where the receiver has access to correlated side information. See~\cref{fig:sys} for an illustration. Specifically, we focus on pairs of \emph{stereo images} with overlapping fields of view as correlated information sources $\xv, \xv_\mathrm{side}$ $\in \II^{C_\mathrm{in} \times W \times H}$, where $W$ and $H$ denote the width and height of the image, while $C_\mathrm{in}$ represents the R, G and B channels for colored images. The transmitter maps the input image into a complex-valued latent vector $\zv =E_{\Thetav} (\xv, \sigma^2)$, where $E_{\Thetav}:\II^{C_\mathrm{in} \times W \times H} \rightarrow \CC^{k}$ is a nonlınear encoding function parameterized by $\Thetav$, and $k$ is the available channel bandwidth.

We impose an average transmission power constraint $\Pavg$ on the transmitted signal $\zv \in \CC^k$:
\begin{align}
\frac{1}{k}  \norm{\zv}_2^2  \leq \Pavg.
\label{eq:power_constraint}
\end{align}

The receiver subsequently receives the noisy latent vector $\yv \in \CC^k$ as 
 $\yv = \zv + \nv$,
where $\nv \in \CC^{k}$ represents the i.i.d. complex Gaussian noise term i.e., $\nv \sim \Cc\Nc(\vec{0}, \sigma^2 \vec{I}_{k})$. We assume that $\sigma^2$ is known at both the transmitter and the receiver.

A nonlinear decoding function  $D_{\Phiv}:\CC^{k} \rightarrow \II^{C_\mathrm{in} \times W \times H}$, parameterized by $\Phiv$, is employed at the receiver, to reconstruct the input image as:
\begin{align*}
           \hat{\vec{x}} = D_{\Phi} (\yv, \xv_{\mathrm{side}},\sigma^2).
\end{align*} 

\noindent We define the bandwidth ratio $\rho$, which characterizes the available channel resources, as:
\begin{align*}
    \rho \triangleq \frac{k}{ C_\mathrm{in} W H} \,\, \si{channel\,symbols \per pixel}.
\end{align*}
We also define the channel $\mathrm{SNR}$ as:
\begin{align}
    \label{eq:snr}
    \mathrm{SNR} \triangleq 10 \logn{10}{\frac{\Pavg}{\sigma^2}} \,\, \si{\decibel}.
\end{align}

Our learning objective is to minimize the average distortion between the original input image $\xv$ at the transmitter and the reconstructed image $\hat{\xv}$ at the receiver, i.e.,
\begin{align*}
    \argmin{\Thetav,\Phiv} \EE \left [ d \left(\xv, \hat{\xv} \right) \right ],
\end{align*}
where the expectation is over the source and side information statistics, $(\xv, \xv_{\mathrm{side}}) \sim p(\xv, \xv_{\mathrm{side}})$, as well as the channel noise distribution. Here, $d (\cdot, \cdot)$ can be any differentiable distortion measure.

\section{Methodology}
\label{sec:methodology}
\begin{figure*}[tbp!]
    \centering
    \includegraphics[width=0.8\linewidth]{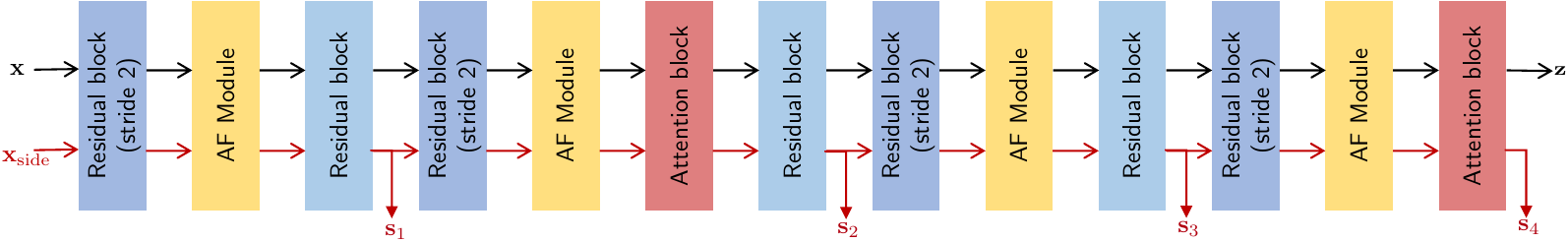}
    \vfill
     \includegraphics[width=0.8\linewidth]{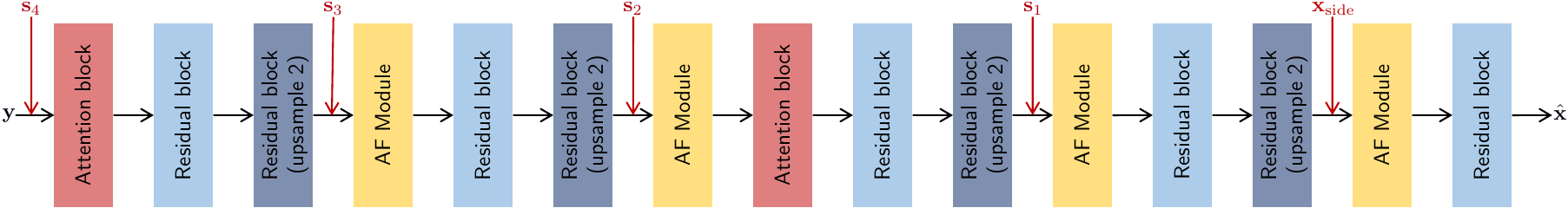}
    \caption{Encoder architecture of our method (\textbf{top}), which is used to encode both the input image $\xv$ at the transmitter and the side information image $\xv_\mathrm{side}$ at the receiver side. Red arrows indicate the flow of the encoded side information. $\sv_1$, $\sv_2$, $\sv_3$ and $\sv_4$ denote the encoded side information at different scales, which are to be used at the receiver side. Decoder architecture of our method (\textbf{bottom}), which is used to reconstruct the input image from the noisy channel output $\yv$ and the side information $\xv_\mathrm{side}$.}
    \label{fig:wz_enc}
\end{figure*}

In this section, we describe our novel architecture that we build upon the current state-of-the-art DeepJSCC architecture. 


\begin{algorithm}[t]
 \scriptsize
   \caption{\strut  Training procedure of DeepJSCC-WZ}
    \label{alg:overview}
    \begin{algorithmic}  
    \Repeat
    \Comment{Iterate through epochs}
        \State Shuffle $\Dctrain$ and set $l=t=0$
        \ForAll{$(\xv,\xv_\mathrm{side}) \in \Dctrain$}
            \LComment{At the transmitter}
            \State Calculate $\sigma^2$ via~\eqref{eq:snr} for $\mathrm{SNR} \sim \mathrm{Uniform}\LSB -5,5 \RSB$
            \State $\tilde{\zv} = E_{\Thetav} \LB \xv, \sigma^2 \RB$ \Comment{Encoding}
            \State $\zv = \sqrt{k \Pavg/\norm{\tilde{\zv}}_2^2} \tilde{\zv}$ \Comment{Induce power normalization}
            \LComment{\gls{AWGN} channel}
            \State Sample $\nv \sim \Cc\Nc(\vec{0}, \sigma^2 \vec{I}_{k})$
            \State $\yv = \zv + \nv$
            \Comment{Received signal}
            \LComment{At the receiver}
            \State $\hat{\xv} = D_{\Phiv} (\yv, \xv_\mathrm{side}, \sigma^2)$
            \Comment{Decoding}
            \State $l = l + \Lc \LB \xv, \hat{\xv} \RB$
            \State t = t + 1
            \If{$t \,\, \mathrm{mod} \,\, batch\, size = 0$}
                \Comment{Standard mini-batch training}
                \State Update $\Thetav, \Phiv$ by backpropagating $l$ 
                \State Reset $l$ and gradients to zero
            \EndIf
        \EndFor
    \Until{validation \gls{PSNR} does not improve for $e$ epochs}
    \end{algorithmic}
\end{algorithm}

\subsection{DeepJSCC-WZ Architecture}\label{subsec:deep_jscc_wz}
Our \gls{DNN} architecture tackles several challenges at once: learning with side information, high dimensional inputs, large number of parameters, spatial correlations, and \gls{SNR} adaptivity. Fig.~\ref{fig:wz_enc} illustrates the encoder and decoder architectures we propose for distributed variant of \gls{DeepJSCC} in the Wyner-Ziv setting, i.e., having decoder-only side information.


We first construct deep \gls{CNN} and autoencoder-based architecture for \emph{DeepJSCC} following the architectures proposed in~\cite{bourtsoulatze2019deep,kurka2020deepjscc}. \glspl{CNN} allow parameter-efficient extraction of high-level features by exploiting spatial structures within the images, and has been known to perform well for various vision-related tasks, including \gls{DeepJSCC} for image transmission~\cite{bourtsoulatze2019deep}. Our encoder and decoder architectures are similar to~\cite{tung2022deepjscc}, which have nearly symmetric encoder and decoder networks. Additionally, they utilize residual connections and a computationally efficient attention mechanism introduced in~\cite{cheng2020learned}. We then adapt \gls{AF} module to enhance the generalization of our model over different \gls{SNR} ranges, without incurring any significant performance degradation. \Gls{AF} module requires \gls{SNR} as an input, which is calculated via~\cref{eq:snr}, in addition to the previous layer's output. Note that we randomly sample \gls{SNR} values during training to enable generalization (see \Cref{alg:overview}).

Reliably incorporating information from multiple views or modalities for learning-based methods, such as \glspl{DNN}, has been known to be challenging. Fusing multiple views can be done at the data, feature or output levels~\cite{mandira2019spatiotemporal,giritliouglu2021multimodal}. In \cite{mital2022neural}, the extracted side information features are concatenated with the received compressed features only before being fed to the decoder. Inspired by these works, we opt for fusing $\xv_\mathrm{side}$ to our decoder at four different stages with different scales, including the encoder's intermediate features and its output. This way, we are able to incorporate both high-level details, such as the presence of objects, from low-resolution features, and fine-grained details, such as the texture of pixels, from the higher resolution features.
    
In the \emph{DeepJSCC-WZ} network, we incorporate the decoder-only side information by employing the same encoder architecture at the receiver, but feed it with $\xv_\mathrm{side}$ as input, instead of the original source image. This encoder's intermediate outputs are given to the decoder at different scales, as shown in Fig.~\ref{fig:wz_enc}. As seen, $\sv_4$ is concatenated at the filter dimension of the noisy channel output $\yv$. Also, $\sv_3, \sv_2$ and $\sv_1$ are concatenated after the consecutive upsampling layers in the decoder. Lastly, the side information $\xv_\mathrm{side}$ is concatenated with the last upsampling layer's output at the filter dimension. 






We satisfy the power constraint in (1) by normalizing the signal at the encoder output $\Tilde{\zv}$ via
\begin{align}
    \zv = \sqrt{k \Pavg} \frac{\Tilde{\zv}}{\sqrt{\Tilde{\zv}^H \Tilde{\zv}}}, 
\end{align}
where $\Tilde{\zv}^H$ refers to the Hermitian transpose of $\Tilde{\zv}$. 





\subsection{Training Loss}

\Cref{alg:overview} shows the overall training procedure DeepJSCC-WZ, which is trained in an unsupervised fashion using only the raw image pairs, $(\xv, \xv_\mathrm{side}) \in \Dctrain$, where $\Dctrain$ is the set of training data pairs. As such, it does not rely on any costly human labelling. We randomly generate and use the channel model throughout training as described in \Cref{subsec:deep_jscc_wz}. We train the whole DNN architecture end-to-end with the task of reconstructing the input image $\xv$ by minimizing the prescribed distortion measure $d(\cdot, \cdot)$. Note that DeepJSCC models can be directly optimized for any differentiable distortion measure, and can be easily adapted to various multi-media transmission tasks (e.g., video as in~\cite{tung2021deepwive}). Following the reasoning in~\cite{mentzer2020highfidelity}, we choose a composite loss function, $\Lc$, to jointly optimize the \emph{distortion-perception trade-off}~\cite{erdemir2022generative,yilmaz2023high} for the reconstructed images as:
\begin{align}
\Lc \LB \xv, \hat{\xv} \RB = \sum_{\substack{\mathclap{(\xv, \xv_\mathrm{side}) \in \Dctrain}}} \mathrm{MSE} \LB \xv, \hat{\xv} \RB + \lambda \cdot \mathrm{LPIPS} \LB \xv, \hat{\xv} \RB ,\label{eq:loss}
\end{align}
where $\mathrm{MSE}(\xv,\hat{\xv}) \triangleq \frac{1}{m} \norm{\xv - \hat{\xv}}_2^2$ is the \gls{MSE} loss, where $m$ is the total number of elements in $\xv$, i.e., $m=C_\mathrm{in} W H$, and \gls{LPIPS}~\cite{zhang2018unreasonable} is a commonly used perceptual quality metric. Here, $\lambda$ is the trade-off parameter, determining the relative importance of \gls{LPIPS} loss w.r.t. the \gls{MSE} loss. Adapted for predicting the similarity of distorted patches, \gls{LPIPS} computes the distance in the feature space of a DNN model that is originally trained for image classification task. Unlike the \gls{MSE} distortion metric (which is applied point-wise), perception-oriented distortion criteria, such as \gls{LPIPS} and \gls{MS-SSIM}~\cite{wang2003multiscale}, measure the similarity between the original image and the reconstructed one using a proxy for the human perception.


\section{Numerical Results and Discussion}
\label{sec:numerical_results}
In this section, we present our experimental setup to show the performance gains of our method in different scenarios.

\subsection{Datasets and Baselines}

For the first part of the experiments, we use the \emph{Cityscape} dataset~\cite{cordts2016cityscapes}, consisting of \num{5000} stereo image pairs, where \num{2975} pairs are used for training, and \num{500} and \num{1525} pairs were used for validation and test, respectively.  For the second part of experiments, following~\cite{ayzik2020deep, mital2022neural}, we compose our dataset from KITTI 2012~\cite{6248074} and KITTI 2015 datasets~\cite{menze_1, menze_2}, termed as \emph{KITTIStereo}, where \num{1576} pairs are used for training, and we validated and tested models on \num{790} image pairs each.



To quantify how much our scheme can benefit from the decoder-only side information, we first compare our method with conventional \emph{DeepJSCC}, which ignores the side information. In addition, we also experiment with further reduction in the number of parameters by using the same set of parameters for encoding both the input image $\xv$ at the transmitter, and $\xv_\mathrm{side}$ at the receiver side, called \emph{DeepJSCC-WZ-sm}. In other words, unlike \emph{DeepJSCC-WZ}, the encoding functions at the transmitter and at the receiver share the same parameters $\Thetav$. DeepJSCC-WZ-\emph{sm} also receives binary indicator of encoding either $\xv$ or $\xv_\mathrm{side}$ to encoder's \gls{AF} modules. For a complete comparison, we further introduce a model named \emph{DeepJSCC-Cond}, where the transmitter \emph{also} has access to $\xv_\mathrm{side}$ image, which serves as an upper bound on the performance of the model of interest. 
We also consider separate source and channel coding designs using the neural image compression approach named \emph{DeepNIC+Capacity}, which incorporates side information as in~\cite{mital2022neural} followed by ideal capacity-achieving channel codes. 
For the performance of DeepNIC+Capacity, we use the results from the original paper~\cite{mital2022neural} and adopt an upper bound for this scheme by equating the reported average rate values in~\cite{mital2022neural} to the capacity of a complex AWGN channel multiplied by the BR value.



\subsection{Implementation and Training Details}

We utilize standard hyperparameters for our method that has been commonly used in the literature~\cite{bourtsoulatze2019deep,tung2022deepjscc}. We employ a learning rate of \num{1e-4}, a batch size of \num{32}, and an average power constraint of $\Pavg=1.0$ (see \Cref{eq:power_constraint}) for all the analyzed methods. We use Adam optimizer~\cite{kingma2014adam} to minimize the training loss in \Cref{eq:loss}. We train the network with channel noise $\nv$ determined by the uniformly sampled SNR between \num{-5} and \num{5} \si{\decibel}. 
Following~\cite{mital2022neural, mital2023neural}, we centre-crop each $375 \times 1242$ image of the KITTIStereo dataset to obtain images of size $370 \times 740$, and then downsample them to  $128 \times 256$. For the Cityscape dataset, we directly downsample each image to the size of $128 \times 256$.



\begin{figure}
    \centering
    \begin{tabular}[b]{ccc}
 \footnotesize{DeepJSCC} & \footnotesize{DeepJSCC-WZ} & \footnotesize{DeepJSCC-Cond} \\
    \begin{subfigure}
        \centering    \includegraphics[width=0.29\linewidth]{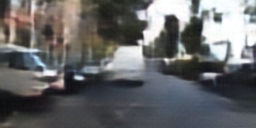}
    \end{subfigure}&
    \begin{subfigure}
        \centering    \includegraphics[width=0.29\linewidth]{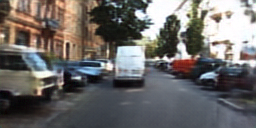}
    \end{subfigure}&
    \begin{subfigure}
        \centering    \includegraphics[width=0.29\linewidth]{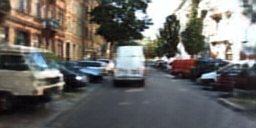}
    \end{subfigure}
    \\
    \begin{subfigure}
        \centering    \includegraphics[width=0.29\linewidth]{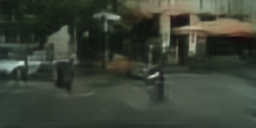}
    \end{subfigure}&
    \begin{subfigure}
        \centering    \includegraphics[width=0.29\linewidth]{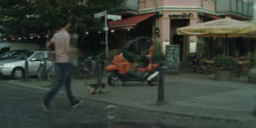}
    \end{subfigure}& 
    \begin{subfigure}
        \centering    \includegraphics[width=0.29\linewidth]{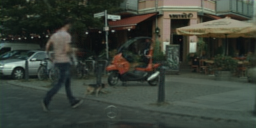}
    \end{subfigure}\\
    \end{tabular}
    \caption{Visual comparison of reconstructed images from KITTIStereo (\textbf{top}) and Cityscape (\textbf{bottom}) datasets, having \gls{BR} value of $\rho={1/32}$ and $\mathrm{SNR}_{\mathrm{test}}=-4$ \si{\decibel}.}
    \label{fig:reconstructed_images}
\end{figure}

\newcommand{\allfigures}[7]{
        \centering
\begin{tikzpicture}[scale=0.63]
        \begin{axis}[
        title={#1 - $\rho \in \{1/16; 1/32\}$},
        xlabel={$\mathrm{SNR}_\mathrm{test}$ (\si{\decibel})},
        error bars/y dir=both,
        error bars/y explicit,
        ylabel={PSNR (\si{\decibel}) $\uparrow$},
        cycle list/Set2-5,
        cycle multiindex* list={
                [5 of]mark list*\nextlist
                Set2-5\nextlist                {solid,solid,solid,solid,solid,dashed,dashed,dashed,dashed,dashed,solid,dashed}\nextlist
        },
        ymin=#2,
        ymax=#3,
        xmin=-5,
        xmax=5
        ]
        \foreach \bwfactor in {16,32}{
        \addplot table[x=model/snr, y=test/psnr, y error=test/psnr_std, col sep=comma]{results/#1_\bwfactor_DeepJSCCWZJoint2.csv};
        \label{label_deepjscc_cond_\bwfactor_#1}
        \addplot table[x=model/snr, y=test/psnr, y error=test/psnr_std, col sep=comma]{results/#1_\bwfactor_DeepJSCCWZBaseline2.csv};
        \label{label_deepjscc_wz_\bwfactor_#1}
        \addplot table[x=model/snr, y=test/psnr, y error=test/psnr_std, col sep=comma]{results/#1_\bwfactor_DeepJSCCWZ.csv};
        \label{label_deepjscc_wz_sm_\bwfactor_#1}
        \addplot table[x=model/snr, y=test/psnr, y error=test/psnr_std, col sep=comma]{results/#1_\bwfactor_DeepJSCCWZBaseline.csv};
        \label{label_deepjscc_\bwfactor_#1}
        \addplot table[x=model/snr, y=test/psnr, col sep=comma]{results/#1_\bwfactor_Separation_psnr.csv};
        \label{label_deepnic_\bwfactor_#1}
        }
        \end{axis}
        \end{tikzpicture}
        \begin{tikzpicture}[scale=0.63]
        \begin{axis}[
        title={#1 - $\rho \in \{1/16; 1/32\}$},
        xlabel={$\mathrm{SNR}_\mathrm{test}$ (\si{\decibel})},
        error bars/y dir=both,
        error bars/y explicit,
        ylabel={MS-SSIM $\uparrow$},
        cycle list/Set2-5,
        cycle multiindex* list={
                [5 of]mark list*\nextlist
                Set2-5\nextlist                {solid,solid,solid,solid,solid,dashed,dashed,dashed,dashed,dashed}\nextlist
        },
        ymin=#4,
        ymax=#5,
        xmin=-5,
        xmax=5
        ]
          \foreach \bwfactor in {16,32}{
        \addplot table[x=model/snr, y=test/msssim, y error=test/msssim_std, col sep=comma]{results/#1_\bwfactor_DeepJSCCWZJoint2.csv};
        \addplot table[x=model/snr, y=test/msssim, y error=test/msssim_std, col sep=comma]{results/#1_\bwfactor_DeepJSCCWZBaseline2.csv};
        \addplot table[x=model/snr, y=test/msssim, y error=test/msssim_std, col sep=comma]{results/#1_\bwfactor_DeepJSCCWZ.csv};
        \addplot table[x=model/snr, y=test/msssim, y error=test/msssim_std, col sep=comma]{results/#1_\bwfactor_DeepJSCCWZBaseline.csv}; 
        \addplot table[x=model/snr, y=test/msssim, col sep=comma]{results/#1_\bwfactor_Separation_msssim.csv};
}
        \end{axis}
        \end{tikzpicture}
        \begin{tikzpicture}[scale=0.63]
        \begin{axis}[
        title={#1 - $\rho \in \{1/16; 1/32\}$},
        xlabel={$\mathrm{SNR}_\mathrm{test}$ (\si{\decibel})},
        ylabel={LPIPS $\downarrow$}, 
        error bars/y dir=both,
        error bars/y explicit,
        cycle list/Set2-5,
        cycle multiindex* list={
                [4 of]mark list*\nextlist
                [4 of]Set2-5\nextlist                {solid,solid,solid,solid,dashed,dashed,dashed,dashed}\nextlist
        },
        ymin=#6,
        ymax=#7,
        xmin=-5,
        xmax=5
        ]
           \foreach \bwfactor in {16,32}{
        \addplot table[x=model/snr, y=test/lpips, y error=test/lpips_std, col sep=comma]{results/#1_\bwfactor_DeepJSCCWZJoint2.csv};
        \addplot table[x=model/snr, y=test/lpips, y error=test/lpips_std, col sep=comma]{results/#1_\bwfactor_DeepJSCCWZBaseline2.csv};
        \addplot table[x=model/snr, y=test/lpips, y error=test/lpips_std, col sep=comma]{results/#1_\bwfactor_DeepJSCCWZ.csv};
        \addplot table[x=model/snr, y=test/lpips, y error=test/lpips_std, col sep=comma]{results/#1_\bwfactor_DeepJSCCWZBaseline.csv};
        }
        \end{axis}
        \end{tikzpicture}
}
\begin{figure*}[!t]
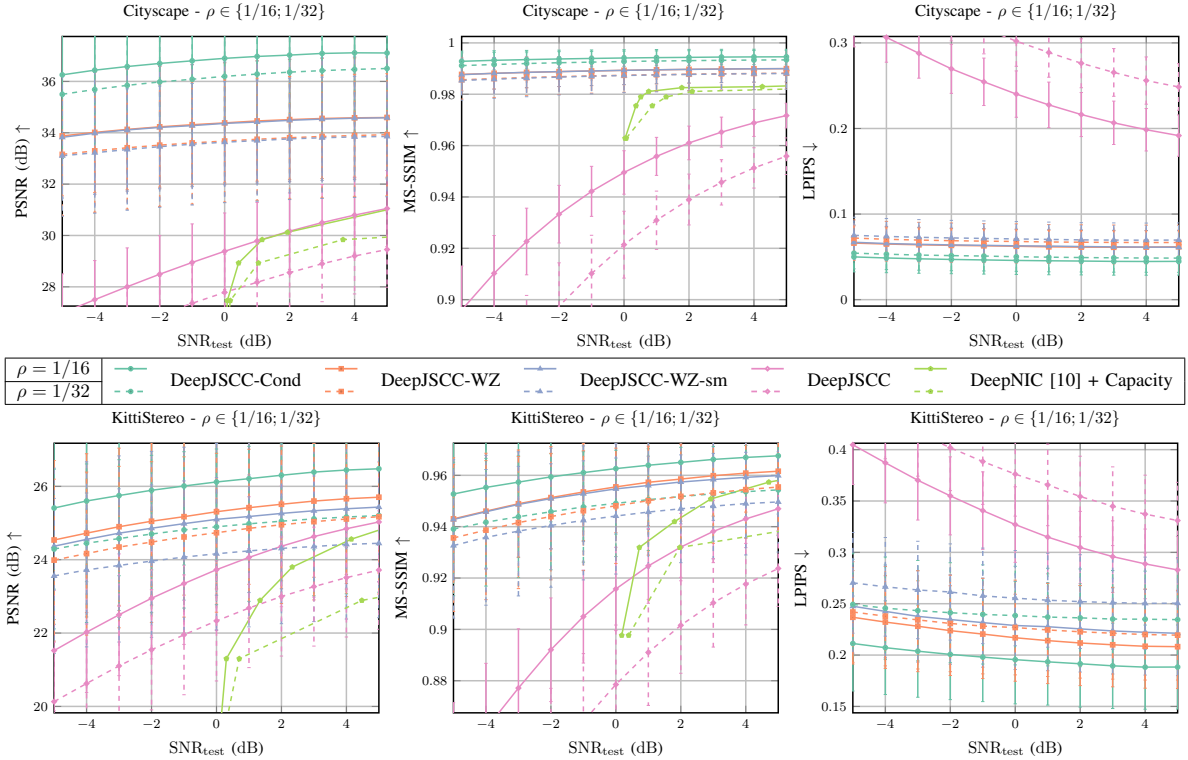

\allfigures{Cityscape}{27.5}{37.5}{0.9}{1.0}{0.0}{0.3}
\resizebox{0.875\textwidth}{!}{
\begin{tabular}{|l|llllllllll|}
\hline
$\rho=1/16$ &
  \ref*{label_deepjscc_cond_16_Cityscape} &
  \multirow{2}{*}{DeepJSCC-Cond} &
  \ref*{label_deepjscc_wz_16_Cityscape} &
  \multirow{2}{*}{DeepJSCC-WZ} &
  \ref*{label_deepjscc_wz_sm_16_Cityscape} &
  \multirow{2}{*}{DeepJSCC-WZ-sm} &
  \ref*{label_deepjscc_16_Cityscape} &
  \multirow{2}{*}{DeepJSCC} &
  \ref*{label_deepnic_16_Cityscape} &
  \multirow{2}{*}{DeepNIC~\cite{mital2022neural} + Capacity} \\ \cline{1-1}
$\rho=1/32$ &
  \ref*{label_deepjscc_cond_32_Cityscape} &
   &
  \ref*{label_deepjscc_wz_32_Cityscape} &
   &
  \ref*{label_deepjscc_wz_sm_32_Cityscape} &
   &
  \ref*{label_deepjscc_32_Cityscape} &
   &
  \ref*{label_deepnic_32_Cityscape} &
   \\ \hline
\end{tabular}
}        
\allfigures{KittiStereo}{20}{27}{0.87}{0.97}{0.15}{0.4}
\caption{Comparison of the introduced DeepJSCC-WZ method with the baselines on the KITTIStereo and Cityscape datasets for \glspl{BR} $\rho \in \{1/16; 1/32\}$.}
\label{fig:comparisons}
\end{figure*}


\subsection{Comparison with the Baselines}

Fig.~\ref{fig:comparisons} demonstrates the performance gains of our proposed method under different \gls{SNR} conditions and \gls{BR} values of $\rho=\{1/16;1/32\}$, on KITTIStereo and Cityscape datasets. We highlight that all the JSCC-based models are trained using the composite loss function in Equation~\eqref{eq:loss}, but are evaluated across various traditional (e.g., \gls{PSNR}) and perceptual (e.g., \gls{MS-SSIM} and \gls{LPIPS}) distortion quality metrics. For both \gls{BR} values, DeepJSCC-WZ outperforms its point-to-point counterpart, that is \gls{DeepJSCC}, as well as its separation-based analogue, that is DeepNIC+Capacity, at all the evaluated \glspl{SNR} in terms of the distortion criteria considered. Unlike the DeepNIC model, we observe that DeepJSCC-WZ does not suffer from the cliff effect and provides a graceful performance degradation as the channel SNR varies. 

Notably, we observe a stark performance improvement in the LPIPS distortion metric, which is widely accepted to be more aligned with human perception of image quality (see Fig.~\ref{fig:reconstructed_images} for some visual examples).

Looking at the performance of DeepJSCC-WZ-sm, we note that imposing the same set of parameters for the encoding of both images, $\xv$ and $\xv_\mathrm{side}$, yields little to no effect on the performance. We also remark that DeepJSCC-WZ achieves a comparable performance with the model DeepJSCC-Joint, whose performance is expected to serve as an upper limit bound on the DeepJSCC-WZ model, considering perceptual distortion criteria such as MS-SSIM and LPIPS. This empirically proves our proposed model's capability of successfully exploiting the decoder-only side information.



\begin{table}[!tbp]
\caption{Number of parameters for the compared methods.}
\centering
\begin{tabular}{@{}lcrr@{}}
\toprule
Method & $\rho=1/16$ & $\rho=1/32$ \\ \midrule
DeepJSCC &   $31.6$M    &     $31.0$M   \\
DeepJSCC-WZ-sm &   $39.8$M	    &   $38.6$M     \\
DeepJSCC-WZ &  $48.9$M	     &  $47.4$M     \\
DeepJSCC-Cond &   $59.1$M    &  $57.4$M    \\
\bottomrule
\end{tabular}
\label{tab:num_parameters}
\end{table}

We refer to Table~\ref{tab:num_parameters} for a comparison of the number of parameters of all the methods we have considered. For the two \gls{BR} values, $\rho=1/16$ and $\rho=1/32$, the amount of increase in the number of parameters is $\approx 25\%$, $\approx 53\%$ and $\approx 85\%$ for DeepJSCC-WZ-sm, Deep-JSCC-WZ and DeepJSCC-Cond, respectively, compared to the standard point-to-point DeepJSCC. DeepJSCC-WZ-sm, DeepJSCC-WZ and DeepJSCC-Cond models have additional filter parameters at the decoder, in comparison to DeepJSCC, in order to fuse the encoded $\xv_\mathrm{side}$ features. DeepJSCC-WZ-sm  has less parameters than DeepJSCC-WZ thanks to the parameter sharing of encoders that encode $\xv$ and $\xv_\mathrm{side}$. Therefore, one can opt for the variant DeepJSCC-WZ-sm, instead of the DeepJSCC-WZ model, for a comparable performance in order to keep the number of parameters within a reasonable budget.  DeepJSCC-Cond has three different encoder modules, two of which are to encode $\xv_\mathrm{side}$ at the transmitter and the receiver, while the other one is to encode $\xv$ at the transmitter. 



\section{Conclusion}

\label{sec:conclusion}


We have introduced a learning-based \gls{JSCC} scheme
for image transmission which is capable of exploiting a decoder-only side information that is in the form of a correlated image. We have demonstrated that the receiver is able to successfully exploit this side information, yielding superior performance over all the channel conditions and distortion criteria considered compared to ignoring the side information. Possible avenues for future work include analyzing the  robustness of the proposed approach considering time-varying channel models and correlation structures for $(\xv, \xv_\mathrm{side})$, and also considering the fully distributed communication scenarios using \gls{DeepJSCC}, which has the potential to be an important ingredient towards realizing the task of practical image/video transmission over wireless sensor networks.


\bibliographystyle{IEEEtran}  
\bibliography{main}


\end{document}